\documentclass[sigconf]{acmart}
\AtBeginDocument{%
  }

\copyrightyear{2025}
\acmYear{2025}
\setcopyright{acmlicensed}\acmConference[CIKM '25]{Proceedings of the 34th ACM International Conference on Information and Knowledge Management}{November 10--14, 2025}{Seoul, Republic of Korea}
\acmBooktitle{Proceedings of the 34th ACM International Conference on Information and Knowledge Management (CIKM '25), November 10--14, 2025, Seoul, Republic of Korea}
\acmDOI{10.1145/3746252.3761108}
\acmISBN{979-8-4007-2040-6/2025/11}
\usepackage{enumitem}
\usepackage{graphicx}
\usepackage{subcaption}
\usepackage{comment}
\usepackage{placeins}

\begin{document}

\title{Reconsidering the Performance of GAE in Link Prediction }

\author{Weishuo Ma}
\affiliation{%
  \institution{Institute for Artificial Intelligence, Peking University}
  \city{Haidian Qu}
  \state{Beijing Shi}
  \country{China}
}
\email{2200013081@stu.pku.edu.cn}

\author{Yanbo Wang}
\affiliation{%
  \institution{Institute for Artificial Intelligence, Peking University}
  \city{Haidian Qu}
  \state{Beijing Shi}
  \country{China}
  }
\email{wangyanbo@stu.pku.edu.cn}

\author{Xiyuan Wang}
\affiliation{%
  \institution{Institute for Artificial Intelligence, Peking University}
  \city{Haidian Qu}
  \state{Beijing Shi}
  \country{China}
  }
\email{wangxiyuan@pku.edu.cn}

\author{Muhan Zhang}
\authornote{Correspondence is to Muhan Zhang}
\affiliation{%
  \institution{Institute for Artificial Intelligence, Peking University}
  \city{Haidian Qu}
  \state{Beijing Shi}
  \country{China}
  }
\email{muhan@pku.edu.cn}


\begin{abstract}
Recent advancements in graph neural networks (GNNs) for link prediction have introduced sophisticated training techniques and model architectures. However, reliance on outdated baselines may exaggerate the benefits of these new approaches. To tackle this issue, we systematically explore Graph Autoencoders (GAEs) by applying model-agnostic tricks in recent methods and tuning hyperparameters. We find that a well-tuned GAE can match the performance of recent sophisticated models while offering superior computational efficiency on widely used link prediction benchmarks. Our approach delivers substantial performance gains on datasets where structural information dominates and feature data is limited. Specifically, our GAE achieves a state-of-the-art (SOTA) Hits@100 score of 78.41\% on the ogbl-ppa dataset. Furthermore, we examine the impact of various tricks to uncover the reasons behind our success and to guide the design of future methods. Our study emphasizes the critical need to update baselines for a more accurate assessment of progress in GNNs for link prediction. Our code is available at \url{https://github.com/GraphPKU/Refined-GAE}.
\end{abstract}




\maketitle

\section{Introduction}

\begin{figure}[t]
\centering
\includegraphics[width=\linewidth]{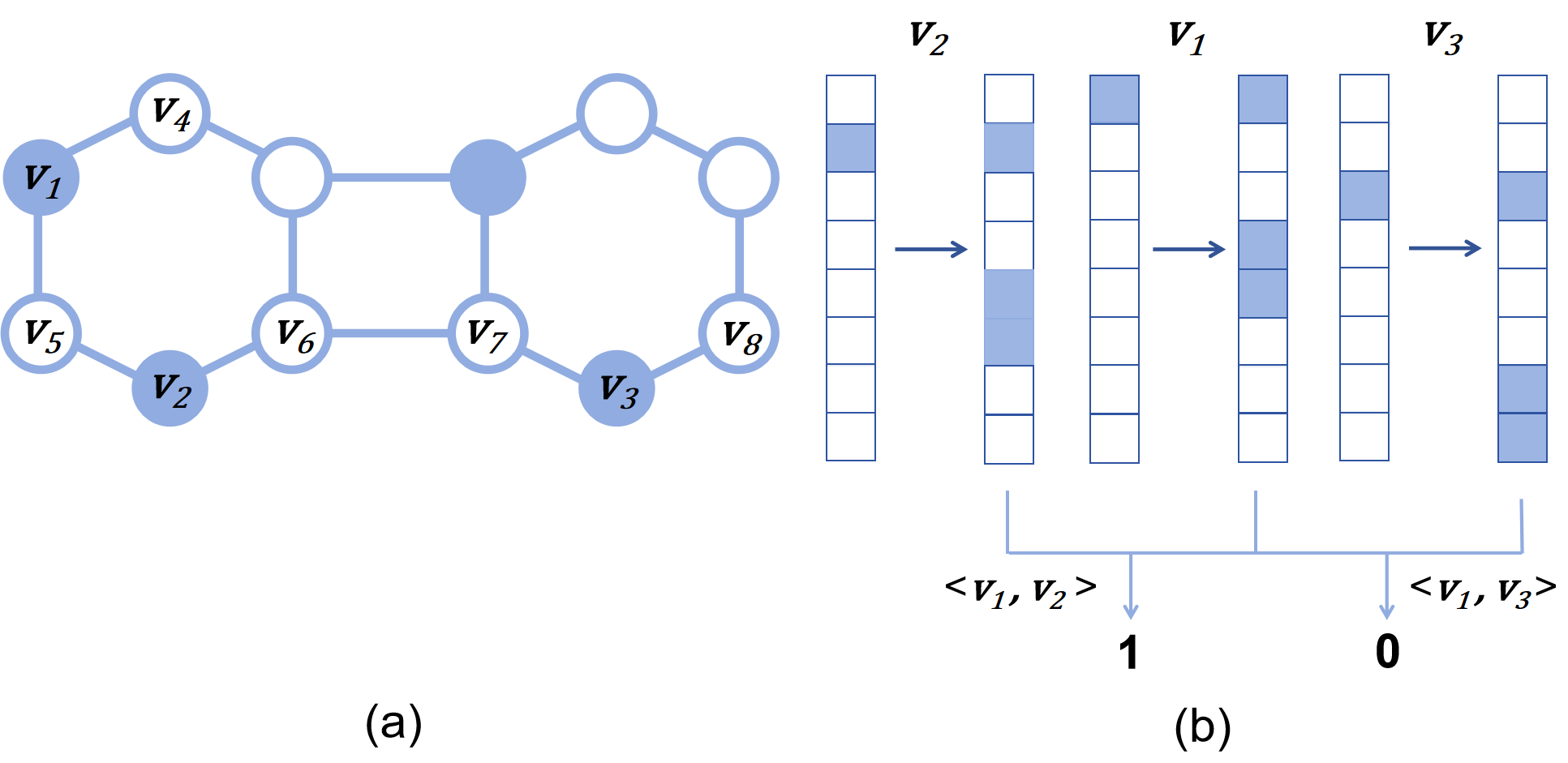} 
\vspace{-2.5em}
  \caption{(a) illustrates an example of the limitations of GAE. Although $v_2$ and $v_3$ share the same representation due to symmetry, the pair $(v_1, v_2)$ should be represented differently from $(v_1, v_3)$. (b) demonstrates how GAE counts common neighbors by performing inner products on the linear combinations of neighboring one-hot embeddings.}
  \label{fig:GAEfailure}
  \Description{The expressiveness limitations of GAE.}
  \vspace{-2em}
\end{figure}

Link prediction is a fundamental problem in graph learning with applications in recommendation systems \citep{recommend}, drug discovery \citep{drug}, and knowledge graphs \citep{kg}. GNNs have achieved strong performance in these domains and are widely adopted for link prediction tasks. Among GNNs, GAE \citep{gae} is a well-known model that predicts link probabilities by computing inner products of node representations learned through a Message Passing Neural Network (MPNN).

However, \citet{seal} found that GAE's expressiveness is limited. For example, as illustrated in Figure~\ref{fig:GAEfailure}a, GAE produces identical predictions for links $(v_1,v_2)$ and $(v_1, v_3)$, despite their differing structural contexts. This limitation arises because GAE computes node representations independently, disregarding the structural relationships between nodes. To address this challenge, various methods have been proposed to enhance expressiveness and better capture graph structure, such as labeling tricks \citep{seal}, paths between nodes \citep{kg}, and neighborhood overlaps \citep{heuristics, neo, buddy, ncn}. These approaches often introduce complex designs with increased time and space complexity, claiming superior performance. Although the expressiveness improvement caused by architecture changes is important, \textbf{we observe that holistic optimization of existing GAE components and meticulous hyperparameter tuning also contribute significantly to these models' success}.

\textbf{This observation motivates a re-evaluation of GAE that moves beyond simple baseline comparisons to instead explore the true potential unlocked when this foundational architecture is enhanced with modern optimization.} We therefore reimplemented GAE, systematically applying principled, relatively simple enhancements and meticulous hyperparameter tuning, practices common in modern GNN. Our approach also introduces a flexible input strategy and key architectural optimizations. These empirical efforts are supported by theoretical analysis; this analysis of GAE with linear propagation and unique inputs reveals an inherent capacity for processing pairwise neighborhood information, akin to capturing generalized common neighbor information. The theoretical grounding, further reinforced by empirical findings such as trained learnable embeddings tending to remain near-orthogonal, validates the effectiveness achieved through optimizing GAE.

Through extensive experiments on Planetoid \citep{planetoid} and Open Graph Benchmark \citep{ogb} datasets, we demonstrate that \textbf{properly Optimized GAE, with its core simplicity largely intact, can match or surpass SOTA models}. As detailed in Table~\ref{table:main_results}, our Optimized GAE achieves \textbf{top-two} performance on all listed datasets and attains the leading average rank compared against all baselines, notably yielding an average improvement of \textbf{6.4\%} over the strongest NCN baseline \citep{ncn}. \textbf{Notably, it establishes a new SOTA Hits@100 score of 78.41\% on the large-scale \texttt{ogbl-ppa} dataset}. These results demonstrate that an optimized GAE model can serve as a powerful and more efficient alternative, comparable to complex GNN architectures that explicitly inject pairwise information. The general applicability of these optimization principles is confirmed by similar performance improvements when applied to NCN \citep{ncn}, highlighting the benefits of a carefully refined base model. Finally, comprehensive ablation studies validate the contribution of each design choice of GAE, providing valuable observations to guide the future development of link prediction models.

In conclusion, we re-evaluate GAE in this study, delivering a stronger baseline for future endeavors. Our main contributions are:

\begin{itemize}[itemsep=2pt,topsep=0pt,parsep=0pt,leftmargin=10pt]
\item We affirm the value of re-evaluating foundational baselines by demonstrating that a GAE, enhanced through principled optimizations, achieves SOTA performance that surpasses many complex models. This result provides a counterpoint to the trend of ever-increasing model complexity, establishing that significant gains can be unlocked by refining simple architectures.
\item We distill our comprehensive analysis into a set of principles for building powerful and efficient GAE-based link prediction models. These guidelines, covering key choices from input representation to architectural configuration, offer an actionable roadmap for the future development.

\end{itemize}

\section{Related Works}

\textbf{GNNs for Link Prediction.} Traditional link prediction methods such as Common Neighbors (CN) \citep{cn}, Adamic-Adar (AA) \citep{aa}, and Resource Allocation (RA) \citep{ra} use fixed heuristics based on shared neighbors. These methods are efficient for capturing local graph structures but are less effective when dealing with complex relationships on diverse datasets. GNNs overcome these issues by learning node representations directly from data. 

The GAE \citep{gae}, an early GNN-based method, uses an MPNN to generate node embeddings and predicts links using inner product similarity. However, its expressiveness is limited because it does not explicitly model pairwise relationships. To address this, SEAL \citep{seal} improves expressiveness by extracting and processing local subgraphs around target links with a GNN. Neo-GNN \citep{neo} and BUDDY \citep{buddy} process the entire graph, using higher-order common neighbor information for improved global modeling. NCN \citep{ncn} further develops pairwise modeling by learning embeddings for common neighbors instead of using manually defined features. Other research has focused on generalizing heuristic methods. NBFNet \citep{kg} unifies path-based heuristics using the Bellman-Ford algorithm, while \citet{mao2023revisiting} categorizes heuristics into local and global ranges for broader structural modeling. LPFormer \citep{shomer2024lpformer} uses an adaptive transformer to dynamically select relevant heuristics for each dataset. Similarly, Link-MOE \citep{mamixture} uses an ensemble of link predictors to flexibly combine multiple heuristics. Another line of research elevates link prediction to a probabilistic generative task, a departure from the methods that explicitly model pairwise relations. Starting with the Variational Graph Autoencoder (VGAE) \citep{gae_variational}, subsequent work has refined this framework, employing methods like incorporating multi-scale information \citep{guo2022multi} or improving generalizability via graph masking \citep{tan2023s2gae}.

Research has also addressed data-related challenges affecting link prediction. For example, LTLP \citep{wang2024optimizing} reduces bias from long-tailed distributions to improve performance on imbalanced datasets. \citet{zhu2024impact} studied the impact of homophily and heterophily. While our study focuses on the static, homogeneous setting, we note that parallel research addresses the challenges in dynamic graphs, such as TGAT \citep{tgat}, and in heterogeneous graphs, such as RGCN \citep{rgcn}.

\noindent\textbf{Baseline Evaluation and Optimization in GNNs.}
As new GNN models are introduced, recent studies suggest caution with evaluation practices that might exaggerate the benefits of complex models. For link prediction, Li et al. \citep{li2024evaluating} identified key problems such as poorly tuned baselines and unrealistic negative sampling in evaluations, which can hide true performance differences between models. Similar concerns exist in other graph tasks. \citet{luo2024a} showed that classic GNNs with carefully tuned under standard methods can perform better than complex Graph Transformers. This applies to node-level \citep{luo2024a} and graph-level tasks \citep{Luo2025a}, questioning the necessity of complex architectures. Complementary to their findings, our work offers a deep dive into link prediction, uncovering specific design choices for GAE beyond general hyperparameter tuning. 

\section{Preliminaries}

\textbf{Notation.}
We consider an undirected graph $G = (V, E, X)$, where $V$ is a set of $n$ nodes, $E$ is the set of edges, and $X \in \mathbb{R}^{n \times d}$ is the $d$-dimensional node feature matrix. Let $A$ be the adjacency matrix. The set of neighbors for node $i$ is $\mathcal{N}(i) = \{ j \in V \mid A_{ij} > 0 \}$. The shortest path distance between $i$ and $j$ is $\text{dist}(i,j)$. The $k$-hop neighborhood of node $i$, $\mathcal{N}_k(i)$, includes all nodes $j$ such that $\text{dist}(i,j) \le k$.

\noindent\textbf{GAE.}
GAEs \citep{gae} are a type of model used for link prediction. GAEs learn low-dimensional node embeddings that preserve graph structure and node attributes. These embeddings are then used to predict link likelihoods. A GAE has two main components:

\noindent\textbf{Encoder:} The encoder uses a MPNN to map node features and graph topology to latent embeddings, which can be formulated as:
\begin{equation}
Z = \text{Encoder}(A, X).
\end{equation}
A common encoder is the Graph Convolutional Network (GCN) \citep{gcn}. Its layer-wise propagation rule is:
\begin{equation}
Z^{(l+1)} = \sigma\left( \tilde{D}^{-1/2} \tilde{A} \tilde{D}^{-1/2} Z^{(l)} W^{(l)} \right),
\end{equation}
where $Z^{(l)}$ is the matrix of node embeddings at layer $l$, $\tilde{A} = A + I_n$ is the adjacency matrix with self-loops, $\tilde{D}$ is the degree matrix of $\tilde{A}$, $W^{(l)}$ is a trainable weight matrix, and $\sigma$ is an activation function.

\noindent\textbf{Decoder:} The decoder uses the latent embeddings $Z$ to reconstruct the adjacency matrix $A$ or predict edge likelihoods. For link prediction, a common approach is to compute the probability of an edge $(i,j)$ using the dot product of their embeddings:
\begin{equation}
\hat{A}_{ij} = \sigma_{sig}\left( Z_i^\top Z_j \right),
\end{equation}
where $\hat{A}_{ij}$ is the predicted probability of an edge, $Z_i$ and $Z_j$ are the latent embeddings of nodes $i$ and $j$, and $\sigma_{sig}$ is the sigmoid function.

The GAE is trained by minimizing an objective function, commonly the binary
cross-entropy (BCE) loss. This loss is calculated over a set of positive
edges $E_{pos}$ (from $E$) and a set of sampled negative edges $E_{neg}$
(node pairs not in $E$). Let $N_{\text{total}} = |E_{pos}| + |E_{neg}|$, and the BCE loss $\mathcal{L}$ is then:
\begin{equation} \label{eq:bce_loss_abbrev}
\mathcal{L} = - \frac{1}{N_{total}} \left( \sum_{(i,j) \in E_{pos}} \log \hat{A}_{ij} + \sum_{(i,k) \in E_{neg}} \log (1 - \hat{A}_{ik}) \right).
\end{equation}

\section{Optimizing GAEs for Link Prediction}
\label{sec4}
The GAE architecture, while a common starting point for link prediction, has well-known limitations in its expressiveness. This section addresses these known constraints by carefully re-evaluating GAE's inherent abilities and showing systematic ways to improve its empirical performance. We begin by theoretically re-examining GAE's core architecture, which reveals an often overlooked ability to process pairwise neighborhood information. Subsequently, using these insights and applying principles of holistic optimization, we introduce the Optimized GAE framework. We detail the systematic design choices and tuning strategies that significantly improve link prediction performance. The specific impact of these design choices is validated in our ablation studies (Section \ref{sec: ablation}).

\subsection{GAE's Inherent Pairwise Structure Modeling}
\label{sec4.1}

To re-examine the GAE architecture for link prediction, it is important to first consider how its expressiveness is typically assessed. Standard analyses often evaluate GNNs, including GAE, by assuming permutation-invariant initial node labels. Under such conditions, these analyses show that GAE's ability to distinguish graphs is similar to the 1-WL test. This highlights a basic limitation in its ability to process purely structural information \citep{seal, morris2019weisfeiler}.

However, GAE applications for link prediction typically use non-permutation-invariant inputs, departing from the assumption of uniform initial labeling. This key difference arises in two main ways: nodes may have unique features, or unique initial embeddings (learnable or distinctly initialized). Either approach gives each node a unique starting signature. This fundamentally changes the GAE's operating conditions compared to those assumed in analyses that rely on initial permutation invariance. Our theoretical exploration starts from this practical consideration.

For analytical clarity, we consider a simple GAE variant. Its encoder has $k$ linear GNN layers that operate on input node features $\mathbf{X} \in \mathbb{R}^{n \times d}$. The resulting node embeddings $\mathbf{Z} \in \mathbb{R}^{n \times d}$ are:
\begin{equation}
    \mathbf{Z} = f_{\text{enc}}(A, \mathbf{X}) = \tilde{A}^k \mathbf{X},
\end{equation}
with $\tilde{A}$ being the normalized adjacency matrix. Using a dot product decoder, the logit for a link $(i,j)$ is $\mathbf{z}_i^T \mathbf{z}_j$. Node $i$'s embedding $\mathbf{z}_i$ is a weighted sum of features from its $k$-hop neighborhood $N_k(i)$:
\begin{equation}
    \mathbf{z}_i = \sum_{p \in {N_k(i)}} \alpha_{ip} \mathbf{x}_p,
\end{equation}
where $\alpha_{ip} = (\tilde{A}^k)_{ip}$ is the weight and $\mathbf{x}_p \in \mathbb{R}^d$ is the initial feature vector of node $p$. Similarly, node $j$'s embedding is $\mathbf{z}_j = \sum_{q \in N_k(j)} \beta_{jq} \mathbf{x}_q$, with $\beta_{jq} = (\tilde{A}^k)_{jq}$. The dot product is:
\begin{align}
\text{logit}(i,j) &= \left(\sum_{p \in {N_k(i)}} \alpha_{ip} \mathbf{x}_p\right)^T \left(\sum_{q \in N_k(j)} \beta_{jq} \mathbf{x}_q\right) \nonumber \\
&= \sum_{p \in {N_k(i)}} \sum_{q \in N_k(j)} \alpha_{ip} \beta_{jq} (\mathbf{x}_p^T \mathbf{x}_q). \label{eq:logit_expansion_multiline}
\end{align}
This expansion shows how the GAE score for a potential link $(i,j)$ is formed. The terms in Equation~\ref{eq:logit_expansion_multiline} where $p=q$ (representing a shared node, which we denote by $r$) sum to $\sum_{r \in N_k(i) \cap N_k(j)} \alpha_{ir} \beta_{jr} ||\mathbf{x}_r||_2^2$. This reflects the influence of these shared nodes $r$, weighted by their initial feature norms. This becomes particularly clear if the initial node embeddings $\mathbf{x}_r$ are orthogonal (e.g., $\mathbf{x}_p^T \mathbf{x}_q \approx 0$ for $p \neq q$) and have approximately unit norm ($\|\mathbf{x}_r\|_2^2 \approx 1$). In such cases, the logit approximately simplifies to $\text{logit}(i,j) \approx \sum_{r \in N_k(i) \cap N_k(j)} \alpha_{ir} \beta_{jr}$. For $k=1$, this sum is $(\tilde{A}^2)_{ij}$, which is a weighted \textbf{common neighbor count} between nodes $i$ and $j$ (since $\alpha_{ir} = (\tilde{A})_{ir}$ and $\beta_{jr} = (\tilde{A})_{jr}$). This way of obtaining common neighbor information from propagated orthogonal inputs is also discussed by \citet{mplp} (illustrated in Figure~\ref{fig:GAEfailure}b). More generally, for propagation depth $k$, the sum $\sum_r (\tilde{A}^k)_{ir} (\tilde{A}^k)_{jr} = (\tilde{A}^{2k})_{ij}$ captures higher-order path overlaps. \textbf{Thus, GAEs, especially with unique orthogonal inputs, inherently capture signals similar to generalized common neighbor information, a key factor for link prediction.}

Furthermore, the terms in Equation~\ref{eq:logit_expansion_multiline} where $p \neq q$ involve dot products $(\mathbf{x}_p^T \mathbf{x}_q)$ between initial features of different nodes $p$ (in $N_k(i)$) and $q$ (in $N_k(j)$). If these initial features $\mathbf{x}_p$ are general (or are learnable embeddings that develop non-zero dot products for $p \neq q$), these terms measure feature similarity between different nodes. For example, strong positive feature correlations between nodes near $i$ and nodes near $j$ might suggest shared community traits. In contrast, feature dissimilarities could indicate contexts less favorable for a link. This ability to assess interactions between different nodes in their neighborhoods allows the model to consider more than just shared nodes, but local environments comprehensively. \textbf{Considering these cross-neighborhood feature similarities is thus crucial, as it allows the model to assess the overall compatibility between the environments of nodes $i$ and $j$.}

\textbf{In conclusion, using unique initial node features or embeddings allows GAEs to inherently perform pairwise information processing.} They can therefore effectively capture common neighbor signals and assess node environment compatibility. This perspective complements traditional analyses of expressiveness that focus on purely structural information.

\subsection{Design Choices for Enhanced GAE}

This section details key architectural design choices for GAEs. The design principles are based on theoretical insights, and their effectiveness is empirically validated in Section \ref{sec: ablation}.

\subsubsection{Input Nodes Representations}
\label{sec:4.2.1}
The insights from Section \ref{sec4.1} guide our approach to designing input node representations. The main decision for initial node representations is whether to use raw node features alone, or to combine them with learned node embeddings. This choice primarily depends on two key factors: first, \textbf{the relative predictive power of inherent node features versus the graph's structure}, and second, \textbf{the overall size and density of the graph}. Carefully evaluating these factors is important for choosing an input strategy that maximizes GAE performance.

To formally measure the relative influence of structure versus features for a dataset $D$ and a link prediction evaluation metric $M$, we consider the performance of simple, deterministic link heuristics. Let $P_S(D, M)$ be the performance of a heuristic that relies \textit{exclusively} on graph structure (e.g., Common Neighbors \citep{cn}). Similarly, let $P_F(D, M)$ be the performance of a heuristic using \textit{only} node features (e.g., the cosine similarity of features). A \textbf{Structure-to-Feature Dominance Index ($I_{S/F}$)} can then be defined as:
\begin{equation}
I_{S/F}(D, M) = \frac{P_S(D, M)}{P_S(D, M) + P_F(D, M) + \epsilon}
\end{equation}
where $\epsilon$ is a small constant to ensure numerical stability if both $P_S$ and $P_F$ are zero. This index, ranging from approximately 0 to 1, offers a principled way to measure the reliance on structure or feature: an $I_{S/F}$ value close to 1 signifies \textbf{structural reliance}, suggesting that topology is the main factor for link formation. Conversely, an $I_{S/F}$ approaching 0 indicates \textbf{feature reliance}. An index around 0.5 suggests a balance, or that neither structure nor features alone are sufficient. \textbf{This $I_{S/F}$ index provides a quantitative basis for tailoring the GAE's input strategy.}

The $I_{S/F}$ index, along with graph size and density, guides the choice of input representation. A low $I_{S/F}$ favors using raw features if available, while a high $I_{S/F}$ suggests learnable embeddings are needed. The suitability of learnable embeddings then depends on the \textbf{number of nodes} ($n$) and the \textbf{average degree}. A higher average degree translates to more per-node connections, enriching the relational data available for training each node's d-dimensional embedding. For large graphs, fixed-capacity encoders and decoders may struggle to adequately represent the vast number of individual nodes, where learnable embeddings become crucial for boosting model capacity. When a graph is both large and dense, the extensive parameters of learnable embeddings are better supported, which helps mitigate the risk of overfitting that can be more pronounced in sparser or smaller graphs. \textbf{Therefore, the optimal input strategy involves a careful synthesis of the $I_{S/F}$ index, the number of nodes, and the average degree.}

When using learnable embeddings, their initialization is an important next step. Orthogonal initialization stands out as a reliable and theoretically sound starting point. As demonstrated in Section 4.1, such initialization allows the GAE's predictive score to directly reflect measures similar to \textbf{weighted common neighbor counts}. By setting the initial dot product between distinct node pairs to zero, this approach creates an unbiased starting point, assuming no arbitrary initial correlations. Later, the model can learn meaningful node embeddings whose pairwise dot products can capture important correlations. \textbf{Therefore, orthogonal initialization for learnable embeddings is very helpful for link prediction.}


\subsubsection{Key Architectural Modules of Graph Autoencoders}

\begin{figure}[t]
\vspace{-1em}
    \centering
    \includegraphics[width=\linewidth]{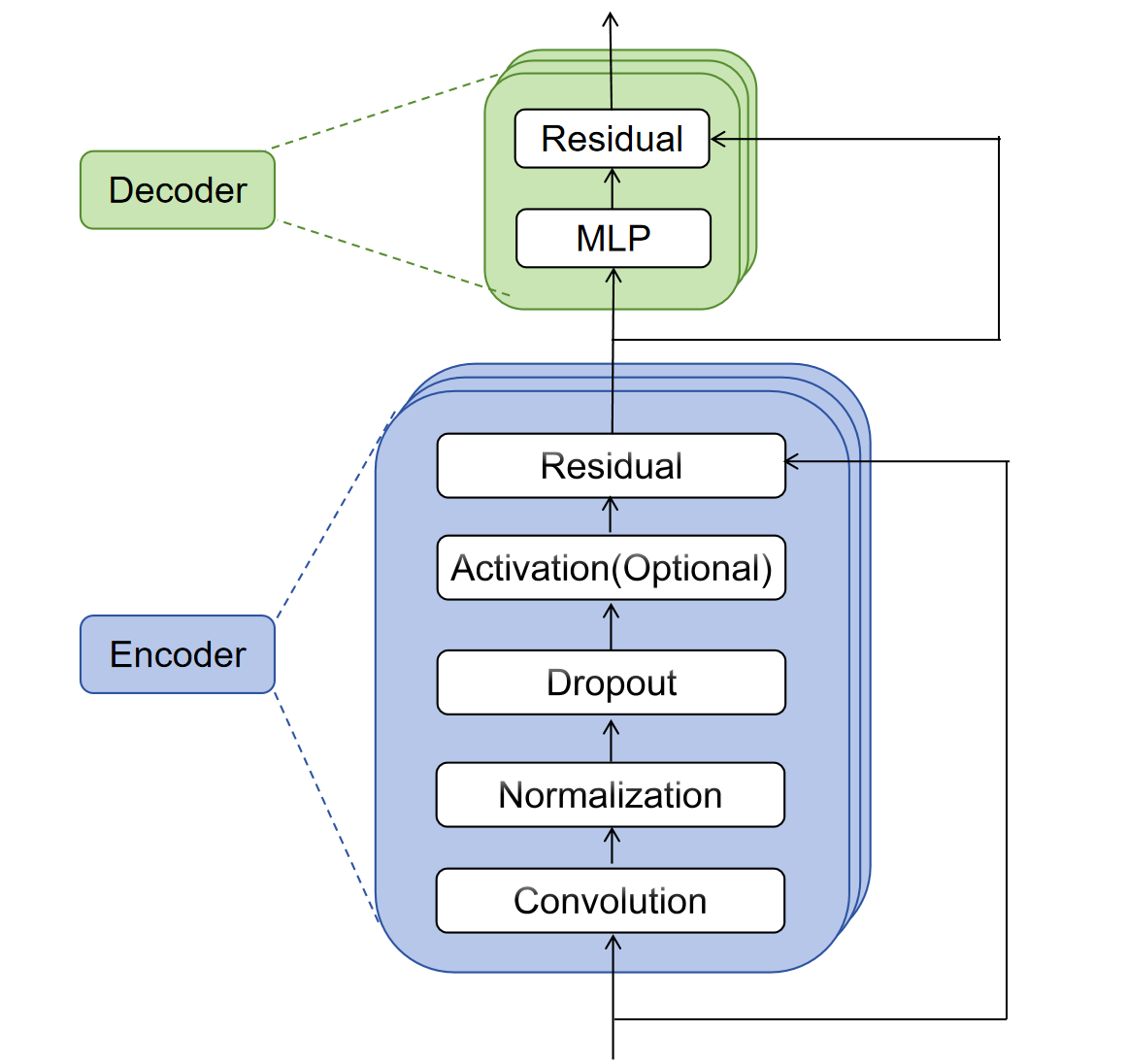}
    \vspace{-2.5em}
    \caption{A visual overview of a typical GAE architecture, highlighting the encoder and decoder modules. }
    \label{fig:archgae}
    \Description{GAE architecture overview.}
\vspace{-2em}
\end{figure}
Following the principled approach for selecting input representations from Section \ref{sec:4.2.1}, we now focus on the GAE's internal architecture. Figure~\ref{fig:archgae} provides a visual overview of this typical architecture. We will now examine the specific components: how the encoder refines embeddings and how the decoder uses them to predict link scores.

The \textbf{encoder}'s role is to transform the input graph structure and the chosen initial node embeddings into lower-dimensional latent embeddings. This is typically achieved using a stack of MPNN layers. Key operations and components in these MPNN layers include:

\begin{table*}[ht]
    \centering
    \caption{Summary of Convolutional Layer Characteristics}
    \vspace{-1em}
    \begin{tabular}{|l|p{4cm}|p{4cm}|p{4cm}|}
        \hline
        \textbf{Convolution Type} & \textbf{Core Characteristics} & \textbf{Benefits} & \textbf{Drawbacks} \\ \hline
        GCN \citep{gcn} & Uses spectral-based normalization for balanced aggregation & Balances node degree differences; robust and reliable & Limited flexibility due to fixed normalization \\ \hline
        GraphSAGE \citep{sage} & Aggregates raw neighbor features using mean aggregator & Simpler aggregation; efficient on certain tasks & Weaker sensitivity to graph structure variations \\ \hline
        GAT \citep{gat} & Employs attention mechanism for weighted aggregation & Effective on graphs with heterogeneous connectivity & High computational cost; prone to overfitting \\ \hline
        GIN \citep{gin} & Incorporates MLP to enhance structural expressiveness & Highly expressive; strong on distinguishing structural details & Higher risk of overfitting, especially on smaller datasets \\ \hline
    \end{tabular}
    \label{tab:conv_summary}
    \vspace{-1em}
\end{table*}

\begin{table*}[t]
\centering
\small
\caption{Hyperparameter Search Ranges and Selected Values for Each Dataset}
\label{table:tbd4}
\begin{tabular}{l c c c c c c c c}
\hline
\textbf{Hyperparameter} & \textbf{Search Range} & \textbf{Cora} & \textbf{Citeseer} & \textbf{Pubmed} & \textbf{ogbl-ddi} & \textbf{ogbl-collab} & \textbf{ogbl-ppa} & \textbf{ogbl-citation2} \\
\hline
Learning Rate (lr)  & $\{5\times10^{-4}, 10^{-4}, 5\times10^{-4}\}$ & $5\times10^{-3}$ & $10^{-3}$ & $10^{-3}$ & $10^{-3}$ & $5\times10^{-4}$ & $5\times10^{-4}$ & $5\times10^{-4}$ \\
MPNN Layers       & $\{2,\,3,\,4\}$        & 4 & 4 & 4 & 2 & 4 & 2 & 3 \\
Hidden Dimension  & $\{256,\,512,\,1024\}$    & 1024 & 1024 & 512 & 1024 & 512 & 512 & 256 \\
Batch Size        & $[2048,\,65536]$     & 2048 & 4096 & 4096 & 8192 & 16384 & 65536 & 65536 \\
Dropout           & $\{0.2,0.4,0.6\}$          & 0.6 & 0.6 & 0.4 & 0.6 & 0.2 & 0.2 & 0.2 \\
Mask Input        & $\{\text{True},\,\text{False}\}$  & True & True & True & True & False & False & False \\
Normalization     & $\{\text{True},\,\text{False}\}$  & True & True & True & False & True & False & True \\
MLP Layers        & $\{2, 4, 5, 8\}$        & 4 & 2 & 2 & 8 & 5 & 5 & 5 \\
\hline
\end{tabular}
\end{table*}

\noindent\textbf{Convolutional Layer.} Convolutional layers are important for determining how node information is aggregated, and their design can significantly affect the model's sensitivity to graph structures. Here, we focus on several commonly used architectures—GCN \citep{gcn}, GraphSAGE \citep{sage}, GAT \citep{gat}, and GIN \citep{gin}—that each use unique methods for feature aggregation. Table \ref{tab:conv_summary} summarizes their different strengths and weaknesses. Among them, GCN is notable for its balanced approach to handling varying node degrees, which makes it both robust and efficient across datasets. \textbf{We use GCN as the primary architecture in our main experiments}.

\noindent\textbf{Residual Connections.} Residual connections \citep{res} are important in deep learning for addressing vanishing gradients and improving training stability. While MPNNs often include self-loops, which offer an implicit way to keep original information, we find it is still helpful to use an \textbf{initial residual}. This design explicitly sums the \textit{original} node features ($Z^{(0)}$) with the output of \textit{each} MPNN layer. Formally, the representation at layer $l$ is computed as:
\begin{equation} \label{eq:initial_residual}
Z^{(l)} = \text{MPNN}_l(Z^{(l-1)}, A) + Z^{(0)}.
\end{equation}
This approach keeps access to the original node attributes throughout the network, preventing them from being diluted by multiple aggregation steps. This strategy is also used in architectures like GCNII \citep{chen2020simpledeep} to help train very deep graph neural networks.

\noindent\textbf{Non-Linear Activation (and its strategic absence).} Section \ref{sec4.1} highlighted that inner products of MPNN-derived embeddings can capture pairwise relationships. However, standard GNN designs often apply non-linear activations after the convolution operation, resulting in updates like $H^{(l+1)} = \sigma(\text{Conv}_{l+1}(H^{(l)}, A))$. Such intermediate non-linearities can disrupt the direct structural information encoded in dot products. \textbf{Therefore, we strategically remove non-linear activations between sequential MPNN layers.} Each layer therefore performs a linear transformation. This approach is similar in principle to linear GNNs like SGC \citep{sgc}, where the update rule is $Z^{(l+1)} = \text{Conv}_{l+1}(Z^{(l)}, A)$. Stacking these linear layers allows the final encoder output to keep structural properties through inner products. The non-linearity is then introduced in the decoder stage. Several other recent works have also effectively used linear propagation for link prediction, including LightGCN \citep{lightgcn}, HL-GNN \citep{hlgnn}, YinYanGNN \citep{yinyang}, and MPLP \citep{mplp}.

After the encoder produces node embeddings, the \textbf{decoder} can use the dot product followed by a sigmoid function to predict edge likelihoods. A more expressive alternative is to use an MLP, which can learn non-linear interactions between node embeddings. This is especially important if the encoder's MPNN layers are linear, as the MLP can then introduce non-linear transformations to model complex relationships. \textbf{Similar to their use in the encoder's MPNN layers, initial residual connections also help in training deeper MLPs in the decoder.} They help ensure stable training and effective information flow through the predictor's layers.

\begin{table*}[t]
    \centering
    \caption{Performance comparison. Metrics include Hits@100, Hits@50, Hits@20, and MRR. The best results are highlighted in bold, and the second-best are underlined. "OOM" indicates Out of Memory. "-" means not reported.}
    \begin{tabular}{lccccccc}
        \toprule
        \textbf{} & \textbf{Cora} & \textbf{Citeseer} & \textbf{Pubmed} & \textbf{Collab} & \textbf{PPA} & \textbf{Citation2} & \textbf{DDI} \\
        \textbf{Metric} & Hits@100 & Hits@100 & Hits@100 & Hits@50 & Hits@100 & MRR & Hits@20 \\
        \midrule
        \textbf{CN}      & $33.92 \pm 0.46$ & $29.79 \pm 0.90$ & $23.13 \pm 0.15$ & $56.44 \pm 0.00$ & $27.65 \pm 0.00$ & $51.47 \pm 0.00$ & $17.73 \pm 0.00$ \\
        \textbf{AA}      & $39.85 \pm 1.34$ & $35.19 \pm 1.33$ & $27.38 \pm 0.11$ & $64.35 \pm 0.00$ & $32.45 \pm 0.00$ & $51.89 \pm 0.00$ & $18.61 \pm 0.00$ \\
        \textbf{RA}      & $41.07 \pm 0.48$ & $33.56 \pm 0.17$ & $27.03 \pm 0.35$ & $64.00 \pm 0.00$ & $49.33 \pm 0.00$ & $51.98 \pm 0.00$ & $27.60 \pm 0.00$ \\
        \midrule
        \textbf{SEAL}    & $81.71 \pm 1.30$ & $83.89 \pm 2.15$ & $75.54 \pm 1.32$ & $64.74 \pm 0.43$ & $48.80 \pm 3.16$ & $87.67 \pm 0.32$ & $30.56 \pm 3.86$ \\
        \textbf{NBFnet}  & $71.65 \pm 2.27$ & $74.07 \pm 1.75$ & $58.73 \pm 1.99$ & OOM & OOM & OOM & $4.00 \pm 0.58$ \\
        \midrule
        \textbf{Neo-GNN} & $80.42 \pm 1.31$ & $84.67 \pm 2.16$ & $73.93 \pm 1.19$ & $57.52 \pm 0.37$ & $49.13 \pm 0.60$ & $87.26 \pm 0.84$ & $63.57 \pm 3.52$ \\
        \textbf{BUDDY}   & $88.00 \pm 0.44$ & \textbf{92.93 $\pm$ 0.27} & $74.10 \pm 0.78$ & $65.94 \pm 0.58$ & $49.85 \pm 0.20$ & $87.56 \pm 0.11$ & $78.51 \pm 1.36$ \\
        \textbf{NCN}     & \textbf{89.05 $\pm$ 0.96} & $91.56 \pm 1.43$ & $\underline{79.05 \pm 1.16}$ & $64.76 \pm 0.87$ & $61.19 \pm 0.85$ & $88.09 \pm 0.06$ & $\underline{82.32 \pm 6.10}$ \\
        \textbf{MPLP+} & - & - & - &\textbf{66.99 $\pm$ 0.40} & $\underline{65.24 \pm 1.50}$ & \textbf{90.72 $\pm$ 0.12} & - \\
        \midrule
        \textbf{GAE(GCN)}     & $66.79 \pm 1.65$ & $67.08 \pm 2.94$ & $53.02 \pm 1.39$ & $47.14 \pm 1.45$ & $18.67 \pm 1.32$ & $84.74 \pm 0.21$ & $37.07 \pm 5.07$\\
        \textbf{GAE(SAGE)}    & $55.02 \pm 4.03$ & $57.01 \pm 3.74$ & $39.66 \pm 0.72$ & $54.63 \pm 1.12$ & $16.55 \pm 2.40$ & $82.60 \pm 0.36$ & $53.90 \pm 4.74$ \\
        \midrule
        \textbf{Optimized GAE} & $\underline{88.17 \pm 0.93}$ & $\underline{92.40 \pm 1.23}$ & \textbf{80.09 $\pm$ 1.72} & $\underline{66.11 \pm 0.35}$ & \textbf{78.41 $\pm$ 0.83} & $\underline{88.74 \pm 0.06}$ & \textbf{94.43 $\pm$ 0.57} \\

        \bottomrule
    \end{tabular}
    \label{table:main_results}
\end{table*}

\begin{table*}[t]
    \centering
    \caption{Comparison of reported performance of NCN and performance after applying our optimization (denoted with *).}
    \begin{tabular}{lccccccc}
        \toprule
         & \textbf{Cora} & \textbf{Citeseer} & \textbf{Pubmed} & \textbf{Collab} & \textbf{PPA} & \textbf{Citation2} & \textbf{DDI} \\
        \textbf{Metric} & Hits@100 & Hits@100 & Hits@100 & Hits@50 & Hits@100 & MRR & Hits@20 \\
        \midrule
        \textbf{NCN} 
         & $89.05 \pm 0.96$ & $91.56 \pm 1.43$ & $79.05 \pm 1.16$ & $64.76 \pm 0.87$ & $61.19 \pm 0.85$ & $88.09 \pm 0.06$ & $82.32 \pm 6.10$ \\
        \textbf{NCN*} 
         & $89.07 \pm 0.34$ & $94.09 \pm 0.40$ & $81.27 \pm 0.57$ & $66.46 \pm 0.67$ & $73.91 \pm 0.44$ & $90.01 \pm 0.08$ & $92.84 \pm 0.70$ \\
         \textbf{Improvement} 
         & (\(\uparrow\) $0.02$) & (\(\uparrow\) $2.53$) & (\(\uparrow\) $2.22$) & (\(\uparrow\) $1.70$) & (\(\uparrow\) $12.72$) & (\(\uparrow\) $1.92$) & (\(\uparrow\) $10.46$) \\
        \midrule
        \textbf{GCN}    
         & $66.79 \pm 1.65$ & $67.08 \pm 2.94$ & $53.02 \pm 1.39$ & $47.14 \pm 1.45$ & $18.67 \pm 1.32$ & $84.74 \pm 0.21$ & $37.07 \pm 5.07$\\
        \textbf{Optimized GAE}
         & $88.17 \pm 0.93$ & $92.40 \pm 1.23$ & $80.09 \pm 1.72$ & $66.11 \pm 0.35$ & $78.41 \pm 0.83$ & $88.74 \pm 0.06$ & $94.43 \pm 0.57$ \\
        \textbf{Improvement} 
         & (\(\uparrow\) 21.38) & (\(\uparrow\) 25.32) & (\(\uparrow\) 27.07) & (\(\uparrow\) 18.97) & (\(\uparrow\) 59.74) & (\(\uparrow\) 4.00) & (\(\uparrow\) 57.36) \\
        \bottomrule
    \end{tabular}
    \label{table:combined_comparison}
\end{table*}

\subsubsection{Revisiting Standard Hyperparameter Settings}
\label{sec:4.2.3}
Beyond the specific architectural components, hyperparameters like \textbf{network depth} and \textbf{hidden dimension} are key in determining a GAE's practical performance. Traditional GAEs often use conservative values for these parameters. This can result in insufficient model capacity, risking underfitting on large graphs and thus underestimating their true potential. Therefore, a careful re-evaluation of standard choices for hidden dimension and network depth is needed to correctly scale the GAE's learning capacity.

\noindent\textbf{Network Depth.} The depth of the network plays an important role in determining its ability to capture complex patterns. However, GNNs have challenges with increasing depth, especially oversmoothing \citep{smooth}. To address the challenge, we adopt the following approach: \textbf{we use shallow MPNN encoder layers yet deep MLP decoder layers.} This design allows the model to retain high expressive power because the MLP learns complex patterns independently.

\noindent\textbf{Hidden Dimension.} A model's capacity is closely linked to its hidden dimension size. However, some previous GNN studies have used relatively small dimensions (e.g., NCN \citep{ncn} using a dimension of only 64 for \texttt{ogbl-collab}). While this reduces computational cost, such limited capacity can greatly underestimate a model's actual performance. \textbf{Therefore, we use a sufficiently large hidden dimension to capture the complexities in large graphs.}

These observations highlight the need to go beyond default or overly conservative settings for network depth and hidden dimension. Careful, dataset-specific tuning of these hyperparameters is important for maximizing the performance of GAEs.

\section{Experiments}
In this section, we present a detailed evaluation of our proposed model. This evaluation has three parts: (1) experimental setup (datasets, baselines, configurations), (2) main results comparing our model against SOTA baselines across multiple datasets, and (3) an ablation study on component contributions and hyperparameter settings. These experiments validate the effectiveness, scalability, and robustness of our approach on several real-world benchmarks.

\subsection{Experiment Setup}
\label{sec:setup}

\subsubsection{Datasets and Baselines}
We evaluate our model on three datasets from Planetoid \citep{planetoid}: \texttt{Cora}, \texttt{Citeseer}, and \texttt{Pubmed}; and four datasets from OGB \citep{ogb}: \texttt{ogbl-collab}, \texttt{ogbl-ddi}, \texttt{ogbl-ppa}, and \texttt{ogbl-citation2}.

As baselines, we use the official OGB results for GAE. For a broader comparison, we also include heuristic methods such as Common Neighbors (CN) \citep{cn}, Adamic-Adar (AA) \citep{aa}, and Resource Allocation (RA) \citep{ra}, as well as GNN models like SEAL \citep{seal}, NeoGNN \citep{neo}, NBFNet \citep{kg}, BUDDY \citep{buddy}, NCN \citep{ncn} and MPLP+ \citep{mplp}. Baseline results are taken from three established studies \citep{ncn, mplp, ogb}.

\subsubsection{Implementation Details}
For the Planetoid datasets, we use a 70\%-10\%-20\% train-validation-test split, while for OGB datasets, we use the official data splits and evaluation metrics. Our main Optimized GAE model incorporates learnable node embeddings for the \texttt{ogbl-ddi}, \texttt{ogbl-ppa}, and \texttt{ogbl-citation2} datasets. For \texttt{ogbl-collab}, we follow the common evaluation strategy of including validation edges only during test time \citep{seal, neo, ogb, buddy, ncn, mplp}.

All experiments are conducted on Nvidia 4090 GPUs, with Nvidia A800 GPUs used for the \texttt{ogbl-citation2} dataset due to its higher computational requirements. We train for up to 500 epochs using a 1:3 negative sampling ratio, and all results are averaged over 5 runs with different random seeds. Hyperparameters were tuned using Bayesian optimization to maximize validation performance, with the final settings for each dataset summarized in Table \ref{table:tbd4}.

\begin{table*}[ht]
\centering
\caption{Dataset Characteristics and Performance Impact of Input Representation Choices. This table details graph statistics and the performance of models configured with additional learnable embeddings (LE) or only raw features (RF). For DDI, which lacks node features, "raw features" are represented by all-ones embeddings. The $I_{S/F}$ index is calculated as $P_S / (P_S + P_F + \epsilon)$, where $P_S$ is the performance of common neighbor heuristic, and $P_F$ represents the performance of cosine similarity heuristic.}
\label{tab:input_impact_summary_transposed}
\begin{tabular}{l|lllllll}
\toprule
\textbf{Characteristic} & \textbf{Cora} & \textbf{Citeseer} & \textbf{Pubmed} & \textbf{Collab} & \textbf{PPA} & \textbf{Citation2} & \textbf{DDI} \\
\midrule
Metric & Hits@100 & Hits@100 & Hits@100 & Hits@50 & Hits@100 & MRR & Hits@20 \\
Perf. with LE & $84.25 \pm 1.29$ & $87.21 \pm 1.23$ & $74.96 \pm 1.69$ & $66.02 \pm 0.12$ & $78.41 \pm 0.83$ & $88.74\pm0.06$ & $94.43 \pm 0.57$ \\
Perf. with RF & $88.17 \pm 0.93$ & $92.40 \pm 1.23$ & $80.09 \pm 1.72$ & $66.11 \pm 0.35$ & $21.37\pm0.16$ & $84.74 \pm 0.21$ & $2.01 \pm 1.14$ \\
Node count & 2,708 & 3,327 & 18,717 & 235,868 & 576,289 & 2,927,963 & 4,267 \\
Avg. Degree & $3.90$ & $ 2.81$ & $ 4.73$ & $5.45$ & $ 52.62$ & $10.44$ & $312.84$ \\
$I_{S/F}$ Index & $ 0.394$ & $ 0.292$ & $ 0.398$ & $ 0.682$ & $ 1.000$ & $ 0.741$ & $ 1.000$ \\
\bottomrule
\end{tabular}
\end{table*}

\begin{table*}[t]
    \centering
    \caption{Ablation study for Optimized GAE, evaluating key design choices  covering: (1) Input Representation; (2) Convolution Layers; (3) Layer Design (impact of residual connections and MPNN encoder linearity).}
    \vspace{-1em}
\begin{tabular}{lccccc}
\toprule
 & \textbf{Cora} & \textbf{Citeseer} & \textbf{Pubmed} & \textbf{DDI} & \textbf{Collab} \\
\textbf{Metric} & Hits@100 & Hits@100 & Hits@100 & Hits@20 & Hits@50 \\
\midrule
\textbf{Optimized GAE} & $88.17 \pm 0.93$ & $92.40 \pm 1.23$ & $80.09 \pm 1.72$ & $94.43 \pm 0.57$ & $66.11 \pm 0.35$ \\
\midrule 
\multicolumn{6}{l}{\textbf{Input Representation}} \\
all-ones learnable embedding & $32.93 \pm 2.73$ & $31.08 \pm 2.52$ & $32.89 \pm 1.23$ & $2.13 \pm 1.46$ & $65.16 \pm 0.14$ \\
random learnable embedding   & $32.12 \pm 2.56$ & $30.99 \pm 3.36$ & $48.72 \pm 2.12$ & $13.84 \pm 1.93$ & $65.77 \pm 0.53$ \\
fixed orthogonal embedding   & $80.02 \pm 2.31$ & $73.81 \pm 0.89$ & $75.88 \pm 0.84$ & $77.61 \pm 1.55$ & $65.88 \pm 0.16$ \\
\midrule
\multicolumn{6}{l}{\textbf{Convolution Layers}} \\ 
SAGE                        & $80.04 \pm 3.40$ & $89.12 \pm 0.40$ & $80.12 \pm 0.43$ & $72.46 \pm 11.23$ & $65.28\pm 0.35$ \\
GAT                         & $80.99 \pm 3.97$ & $87.71 \pm 2.29$ & $81.93 \pm 1.49$ & $89.45 \pm 2.01$ & $66.01 \pm 0.11$ \\
GIN                         & $ 82.52\pm 0.90$ & $68.02 \pm 1.83$ & $83.00 \pm 0.60$ & $96.45 \pm 0.42$ & $67.94 \pm 0.43$ \\
\midrule
\multicolumn{6}{l}{\textbf{Layer Design: Residuals \& Linearity}} \\ 
No Residual in MPNN         & $69.95 \pm 5.77$ & $79.52 \pm 2.48$ & $78.46 \pm 1.39$ & $71.12 \pm 1.52$ & $66.11 \pm 0.16$ \\
No Residual in MLP          & $84.45 \pm 2.29$ & $91.21 \pm 1.23$ & $80.17 \pm 1.88$ & $94.02 \pm 1.05$ & $65.46 \pm 1.34$ \\
No Residual in Both         & $29.10 \pm 18.41$ & $78.51 \pm 1.12$ & $ 78.10 \pm 1.09$ & $26.4 \pm 2.59$ & $65.46 \pm 1.34$ \\
Nonlinear MPNN Layers       & $83.47 \pm 0.77$ & $67.15 \pm 1.97$ & $80.56 \pm 1.33$ & $66.77 \pm 3.45$ & $65.68 \pm 0.63$ \\ 
\bottomrule
\end{tabular}
\label{tab:abl}
\end{table*} 

\subsection{Main Results}

The main results are presented in Table \ref{table:main_results}. Our model achieves SOTA on 3 datasets and top-2 on 4 datasets. The consistently strong performance reflects an average improvement of \textbf{91.5\%} over the original GAE baseline. For instance, on the structure-reliant \texttt{ogbl-ppa} and \texttt{ogbl-ddi} datasets, where our input representation technique is particularly effective, performance improves from 18.67\% to \textbf{78.41\%} and from 37.07\% to \textbf{94.43\%}. Furthermore, Optimized GAE shows an average improvement of \textbf{6.4\%} over NCN \citep{ncn}, a competitive SOTA model, surpassing it on six out of seven datasets. Notably, on \texttt{ogbl-ppa}, our model's Hits@100 score of \textbf{78.41\%} not only exceeds the strong MPLP+ baseline by a significant \textbf{20.2\%} margin but also outperforms the OGB leaderboard leader by \textbf{2.4\%}. These results support our main claim: a properly optimized GAE can match or exceed the performance of current models that use more complex methods, validating GAE's potential in current link prediction tasks.

Additionally, as many recent models use GAE as their foundation, our optimization techniques can also be applied to improve their performance. Table \ref{table:combined_comparison} shows the results for the NCN model when improved with our techniques, \textbf{demonstrating a clear and consistent improvement in its performance}. This highlights the broader impact of our methods for improving general GNNs.

\subsection{Ablation Study}
\label{sec: ablation}

In this section, we conduct a comprehensive ablation study to validate the techniques described in Section \ref{sec4}. This study highlights the importance of the proposed techniques and offers a deeper understanding of how they enhance the model’s effectiveness.

\subsubsection{Input Nodes Representations}
The choice of input representations is fundamental to the Optimized GAE. This involves selecting between raw node features and learnable embeddings based on dataset characteristics, and ensuring that learnable embeddings are well-initialized. The following observations support these ideas:

\vspace{3pt}
\noindent\textbf{Observation 1: The choice to use raw features (RF) alone or augment with learnable embeddings (LE) is mainly guided by the informativeness of raw features relative to the structure.}
\vspace{3pt}

\noindent\textbf{Explanation:} Table~\ref{tab:input_impact_summary_transposed} clearly supports this. A higher $I_{S/F}$ index typically indicates that LEs are more beneficial. For instance, on the \texttt{ogbl-ppa} dataset, which lacks informative raw features ($I_{S/F} \approx 1.000$), using LE substantially improves performance from approximately $21.37$ to $78.41$. Conversely, for a feature-rich dataset like \texttt{Citeseer} ($I_{S/F} \approx 0.292$), using its strong RF alone gives a result of $92.40$, which is better than the $87.21$ achieved if LEs are used. While the overall success of LE also depends on graph size and density (see Observation 2), the $I_{S/F}$ index provides an important first assessment. It helps determine if the node features are sufficient or if the model needs to focus more on the graph structure.

\vspace{3pt}
\noindent\textbf{Observation 2: Besides the $I_{S/F}$ index, graph size (number of nodes) and density (average degree) are key factors in determining whether LE can be beneficial.}
\vspace{3pt}

\noindent\textbf{Explanation:}
The effect of graph size and density on LE effectiveness, even with similar feature informativeness, is clear when comparing \texttt{ogbl-collab} and \texttt{ogbl-citation2}. While both datasets exhibit similar $I_{S/F}$ indices, \texttt{ogbl-citation2} is significantly larger and denser than \texttt{ogbl-collab}. On \texttt{ogbl-citation2}, LEs offer a clear advantage over RF. However, on the smaller and sparser \texttt{ogbl-collab}, RF performance is similar to that of LE. This suggests that larger graph scale and higher density better support LE.

\vspace{3pt}
\noindent\textbf{Observation 3: Orthogonal initialization of learnable node embeddings is a significantly more effective starting point compared to arbitrary initializations.}
\vspace{3pt}

\noindent\textbf{Explanation}:
The major impact of initialization is clear from Table~\ref{tab:abl}. For instance, while Optimized GAE uses LEs with an orthogonal initialization and achieves a Hits@20 score of $94.43$ on \texttt{ogbl-ddi}, performance drops sharply with less structured initializations. Specifically, using an ``all-ones initialization,'' where initial embedding vectors are all ones, results in a score of only $2.13$. Similarly, using ``random initialization,'' where embeddings are initialized with values selected uniformly from $[-1, 1]$ (a method likely to break any initial orthogonality), greatly reduces performance to $13.84$. This large drop in performance highlights the importance of orthogonal initialization. Notably, even after training, these learnable embeddings often remain close to orthogonal. Specifically, on \texttt{ogbl-ddi}, the average absolute cosine similarity was 0.03 (std. dev. 0.04) before training and 0.07 (std. dev. 0.07) after training. This reflects the important role of common neighbor information in link prediction, as discussed in Section~\ref{sec4.1}.

\vspace{3pt}
\noindent\textbf{Observation 4: The adaptability offered by making node embeddings \textit{learnable} is important for achieving best performance, and is better than using fixed embedding strategies.}
\vspace{1pt}

\noindent\textbf{Explanation}:
Table~\ref{tab:abl} also shows the benefits of learnability. While using 'fixed orthogonal embeddings' (initialized orthogonally but not updated during training) on \texttt{ogbl-ddi} achieves a Hits@20 score of $77.61$, this is much lower than the $94.43$ achieved by the full Optimized GAE with learnable embeddings. This performance gap highlights that allowing the embeddings to be fine-tuned enables the model to use more than just common neighbor information and capture detailed representations specific to the graph structure.
\subsubsection{Key Architectural Modules of Graph Autoencoders}
\label{sec:ablation_arch_modules}

Beyond input representation, the design of the GAE's architectural components critically influences its performance. These include the choice of graph convolutional operation, the strategy for incorporating residual connections, and the handling of non-linearities within the network. The following observations, based on the ablation results in Table~\ref{tab:abl}, validate our design principles for these modules.

\vspace{3pt}
\noindent\textbf{Observation 5: While GCN serve as a effective default for the Optimized GAE, alternative convolutional architectures exhibit varying performance trade-offs, underscoring the importance of layer selection tailored to dataset characteristics.}
\vspace{3pt}

\noindent\textbf{Explanation}:
Table~\ref{tab:abl}(section ``Convolution Layers'') compares the performance of Optimized-GAE (which uses GCN by default) against variants employing SAGE, GAT, and GIN. While GCN provides strong performance across datasets, alternatives show varied results. For example, GIN can achieve higher performance on large scale benchmarks like \texttt{ogbl-ddi} ($96.45$) by potentially capturing finer-grained structural details, but it may overfit on small graphs like \texttt{Citeseer} ($68.02$). SAGE often trails GCN, particularly on datasets where degree normalization is beneficial. GAT also generally performs slightly below GCN. These findings affirm GCN as a generally well-performing choice, while highlighting that specific dataset properties might favor other architectures, emphasizing the need for dataset-specific tuning if deviating from GCN.

\begin{figure}[t]
\includegraphics[width=\linewidth]{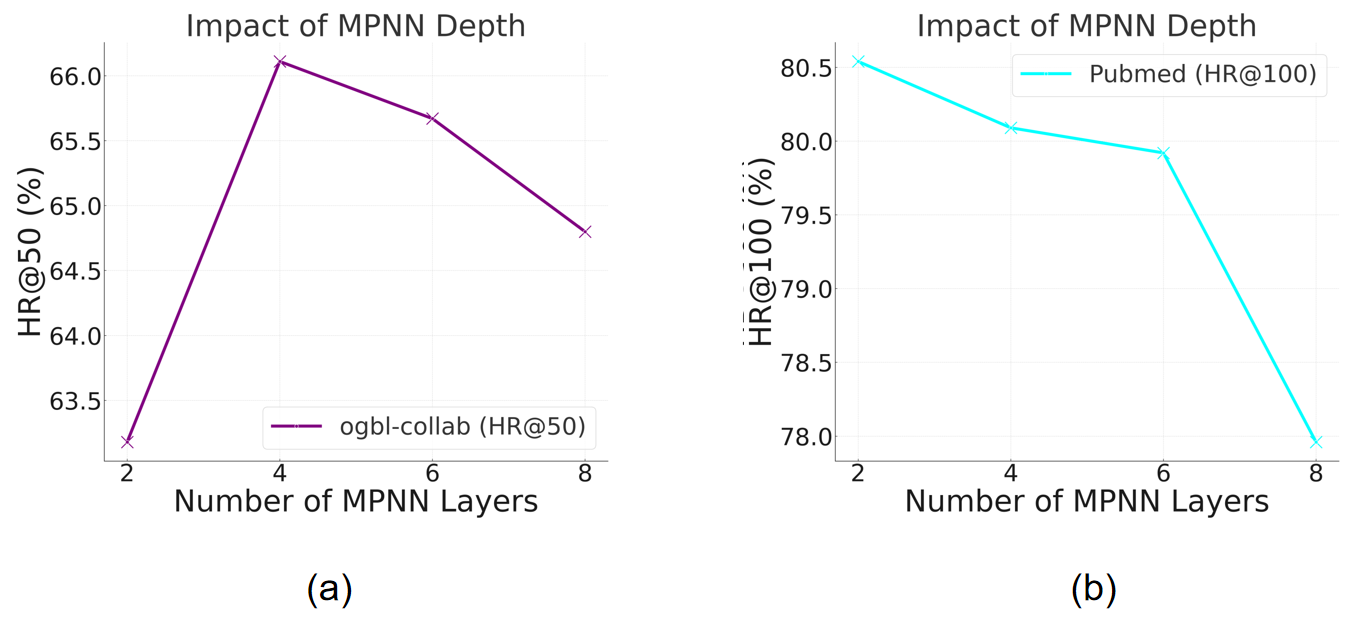} 
\vspace{-2em}
  \caption{Impact of MPNN layer depth on performance for (a) ogbl-collab and (b) Pubmed.}
  \Description{The influence of MPNN layer depth on performance.}
  \label{fig:mpnn}
  \vspace{-1em}
\end{figure}

\begin{figure}[t]
\includegraphics[width=\linewidth]{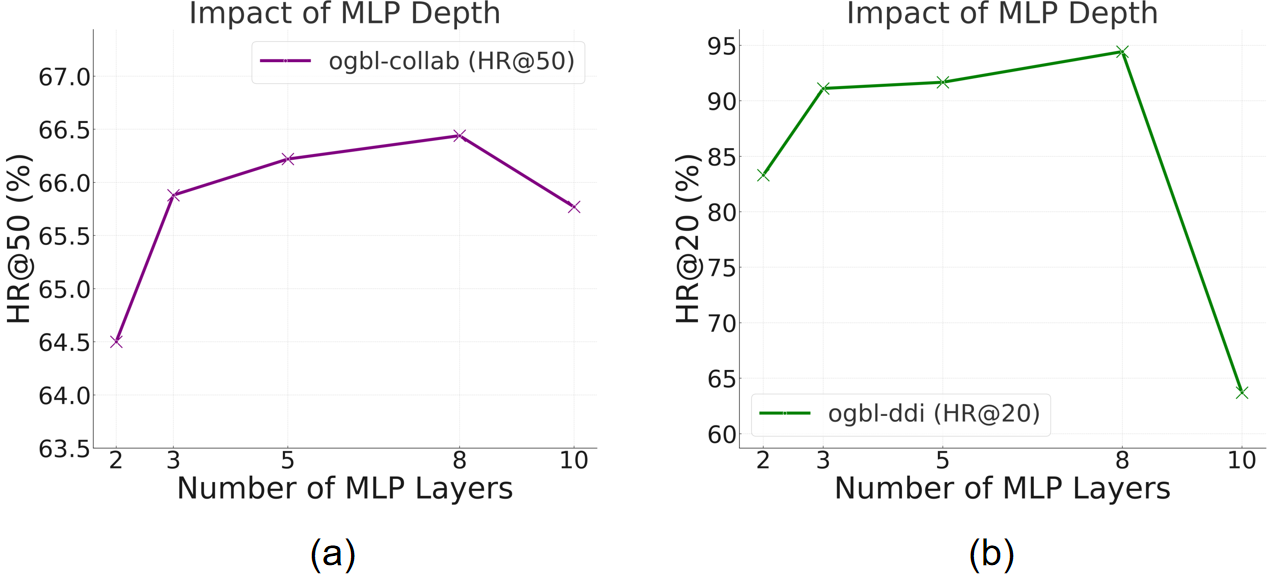} 
\vspace{-2em}
  \caption{Impact of MLP layer depth on performance for (a) ogbl-collab and (b) ogbl-ddi.}
  \Description{The influence of MLP layer depth on performance.}
  \label{fig:mlp}
  \vspace{-1em}
\end{figure}
\begin{figure}[t]
\includegraphics[width=\linewidth]{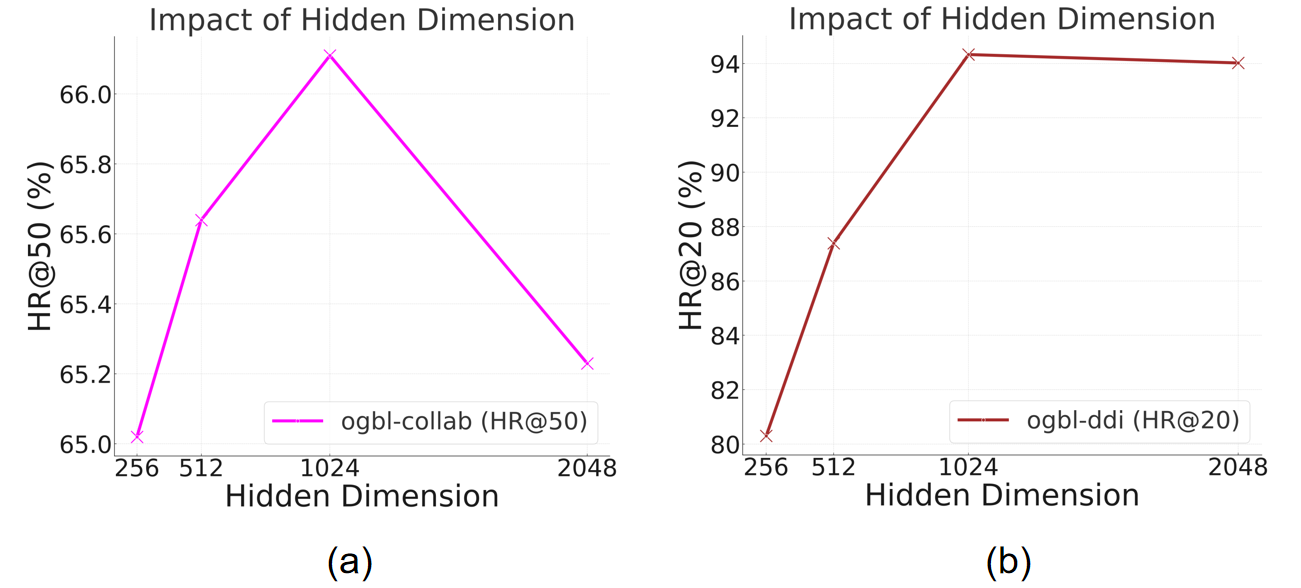} 
\vspace{-2em}
  \caption{Impact of hidden dimension on performance for (a) ogbl-collab and (b) ogbl-ddi.}
  \Description{The influence of hidden dimension on performance.}
  \label{fig:hidden}
  \vspace{-1em}
\end{figure}

\vspace{3pt}
\noindent\textbf{Observation 6: Using initial residual connections in both the MPNN encoder and the MLP decoder is key for stable training and effective information propagation, significantly improving GAE performance.}
\vspace{3pt}

\noindent\textbf{Explanation}:
The role of residual connections is confirmed by ablations in Table~\ref{tab:abl}(section ``Layer Design``). Removing initial residual connections from the encoder causes a substantial performance drop on all datasets. This shows their importance for keeping initial node information during the encoding process. While removing residuals from the decoder has a less dramatic impact on some datasets, it can still harm results (e.g., \texttt{Cora} performance drops to $84.45$). Importantly, removing residual connections from both components results in a severe drop in performance (e.g., \texttt{ogbl-ddi} to $26.4$). Together, these findings highlight that residual connections are necessary for high-performing GAE models.

\vspace{3pt}
\noindent\textbf{Observation 7: Removing non-linear activation functions from the intermediate layers of the MPNN encoder to keep linear message propagation until the decoder stage—is a key factor for preserving structural information and achieving better link prediction performance.}
\vspace{3pt}

\noindent\textbf{Explanation}:
The significant advantage of using linear encoder layers is supported by the ``Nonlinear MPNN Layers'' ablation presented in Table~\ref{tab:abl} (section ``Layer Design''). When non-linear activation functions are added back between the MPNN layers in the Optimized GAE's encoder, performance consistently and often substantially drops on multiple datasets. For example, on \texttt{ogbl-ddi}, the result falls from $94.43$ to $66.77$. These results strongly support our design choice of using linear propagation in the encoder. This allows the model to effectively capture and keep pairwise relational signals and structural information, as discussed in Section~\ref{sec4.1}.

\subsubsection{Hyperparameter Tuning: Network Depth and Hidden Dimension}
\label{sec:ablation_hyperparams}

Beyond architectural choices, the careful tuning of hyperparameters like network depth and hidden dimension, as emphasized in Section~\ref{sec:4.2.3}, is key for achieving the GAE's full potential. The following observations support our strategies for these settings:

\begin{table}[t]
    \centering
    \caption{Comparison of time complexity per training batch and total runtime (seconds) on evaluation sets across various models and datasets. Here, $B$ denotes batch size, and $h$ the hashing sketch size for BUDDY and MPLP+.}
    \label{tab:time}
    \footnotesize
    \begin{tabular}{lccc}
        \toprule
        \textbf{Model} & \textbf{Complexity} & \textbf{ddi} & \textbf{collab} \\
        \midrule
        Optimized GAE      & $O(|E|d + Bd^2)$     & 0.12 & 0.07                       \\
        SEAL            & $O(B|E|d+B|V|d^2)$     & 209 & 1080                      \\
        BUDDY           & $O(|E|d + |E|h + Bd^2 + Bh)$    & 57      & 3.3                \\
        MPLP+           & $O(|E|d + |E|h + |V|d^2 + Bd^2)$    & 0.19     & 0.20                  \\
        \bottomrule
    \end{tabular}
    \vspace{-1em}
\end{table}

\vspace{3pt}
\noindent\textbf{Observation 8: Optimal GAE performance is achieved through a strategy that employs shallow MPNN encoder layers to mitigate oversmoothing, complemented by deeper MLP decoder layers which enhance model capacity.}
\vspace{3pt}

\noindent\textbf{Explanation}:
The impact of MPNN depth is illustrated in Figure~\ref{fig:mpnn}. Performance generally peaks when using 2-4 MPNN layers, with further increases often leading to diminishing returns or a decline, likely due to oversmoothing. This supports our approach of using a relatively shallow MPNN encoder. Conversely, as shown in Figure~\ref{fig:mlp}, increasing the depth of the MLP decoder layers consistently improves performance up to an optimal point where the model sufficiently captures the data complexity. This presents the benefit of using deeper MLPs in the decoder to model complex interactions.

\vspace{3pt}
\noindent\textbf{Observation 9: Using an appropriately sized hidden dimension is important for model capacity; GAE performance generally improves with larger dimensions until the benefits lessen or computational trade-offs become too large.}
\vspace{3pt}

\noindent\textbf{Explanation}:
Figure~\ref{fig:hidden} shows the effect of varying the hidden dimension. Performance typically increases as $d$ is enlarged, allowing the model to represent more complex patterns. For example, on \texttt{ogbl-ddi}, Hits@20 significantly improves when moving from smaller dimensions to $d=1024$. However, beyond a certain size (e.g., $d=1024$ for \texttt{ogbl-collab}), the gains may lessen or even decrease. This highlights the need to choose a suitable hidden dimension that balances capacity with efficiency and helps prevent overfitting from too many parameters. 


\vspace{-0.5em}

\section{Efficiency}

\textbf{In addition to its strong performance, Optimized GAE is also highly scalable due to its simple architecture.} Models like GAE do not require explicit calculation of pairwise information, which gives them an inherent efficiency advantage over models that rely on more complex mechanisms.

The theoretical basis for this efficiency is clear from the model's time complexity. The Optimized GAE encoder uses linear layers, resulting in a complexity of just $O(|E|d)$, while the MLP decoder operates on a batch of $B$ edges with complexity $O(Bld^2)$. This is significantly more efficient than competing methods. For instance, SEAL requires extracting and processing a unique subgraph for each link, leading to a high batch complexity of $O(B|E|d+B|V|d^{2})$. BUDDY must compute hashing sketches over the graph, giving it a complexity of $O(|E|d+|E|h)$ followed by a predictor of $O(Bd^{2}+Bh)$, while MPLP+ adds an extra $O(|E|h)$ for propagating orthogonal sketches on top of its GAE-style backbone. Table \ref{tab:time} compares the theoretical time complexity per training batch and the actual runtime on our evaluation sets. These results highlight that our GAE-based approach offers substantial scalability benefits over models with complex pairwise methods, even when using a large hidden dimension and deeper layers.

\vspace{-0.5em}

\section{CONCLUSION}

This work demonstrates that a holistically optimized GAE can achieve SOTA link prediction performance, rivaling sophisticated contemporary models often with superior efficiency. Our success, achieved through systematic and validated enhancements to GAE's input representation strategies, core architectural designs, and meticulous hyperparameter tuning, underscores the critical importance of re-evaluating simple baseline models. This study not only provides a robustly tuned baseline and practical design guidelines but more broadly highlights that advancements in graph learning can significantly benefit from the thorough understanding and refinement of well-established architectures. Looking ahead, the power and efficiency of our Optimized GAE makes it a strong candidate for future extension to more complex domains such as dynamic and heterogeneous graphs.

\section*{Acknowledgments}
This work is supported by the National Key R\&D Program of China (2022ZD0160300) and National Natural Science Foundation of China (62276003).

\clearpage
\newpage

\section*{GenAI Usage Disclosure}
Generative AI (GenAI) tools were utilized in the preparation of this manuscript to assist with coding and language polishing. The core research ideas, experimental design, analysis, and the overall drafting and writing of the paper were conducted by the authors without the involvement of GenAI.

\bibliographystyle{ACM-Reference-Format}
\bibliography{sample-base}


\end{document}